\documentclass[twoside,leqno,twocolumn]{article}

\usepackage[letterpaper]{geometry}

\usepackage{siamproceedings}

\usepackage[T1]{fontenc}
\usepackage{amsfonts}
\usepackage{graphicx}
\usepackage{epstopdf}
\usepackage{enumitem}
\ifpdf
  \DeclareGraphicsExtensions{.eps,.pdf,.png,.jpg}
\else
  \DeclareGraphicsExtensions{.eps}
\fi

\newsiamremark{remark}{Remark}
\newsiamremark{hypothesis}{Hypothesis}
\crefname{hypothesis}{Hypothesis}{Hypotheses}
\newsiamthm{claim}{Claim}

\usepackage{amsopn}

\usepackage{float}
\usepackage{anyfontsize}
\usepackage{subcaption}

\usepackage{algorithm}
\usepackage{algorithmicx}
\usepackage{algpseudocode}

\usepackage{booktabs}

\usepackage{paralist}
\def\ours{AutoNFS}

\def\R{\mathbb{R}}

\def\loss{\mathcal{L}}
\def\B{\mathcal{B}}
\begin{document}

\newcommand\relatedversion{}
\renewcommand\relatedversion{\thanks{The full version of the paper can be accessed at \protect\url{https://arxiv.org/abs/0000.00000}}} % Replace URL with link to full paper or comment out this line

\title{AutoNFS: Automatic Neural Feature Selection}
    % \author{Anonymous Authors}
    
    \author{Witold Wydma\'nski
    \and Marek \'Smieja\thanks{Faculty of Mathematics and Computer Science, Jagiellonian University, Krak\'ow, Poland
  (\email{wwydmanski@gmail.com}, \email{marek.smieja@uj.edu.pl}).}}

\date{}

\maketitle

%\pagenumbering{arabic}
%\setcounter{page}{1}%Leave this line commented out.

\begin{abstract}
Feature selection (FS) is a fundamental challenge in machine learning, particularly for high-dimensional tabular data, where interpretability and computational efficiency are critical. Existing FS methods often cannot automatically detect the number of attributes required to solve a given task and involve user intervention or model retraining with different feature budgets. Additionally, they either neglect feature relationships (filter methods) or require time-consuming optimization (wrapper methods). To address these limitations, we propose \ours{}, which combines the FS module based on Gumbel-Sigmoid sampling with a predictive model evaluating the relevance of the selected attributes. The model is trained end-to-end using a differentiable loss and automatically determines the minimal set of features essential to solve a given downstream task. Unlike many wrapper-style approaches, \ours{} introduces a low and predictable training overhead and avoids repeated model retraining across feature budgets. In practice, the additional cost of the masking module is largely independent of the number of input features (beyond the unavoidable cost of processing the input itself), making the method scalable to high-dimensional tabular data. We evaluate \ours{} on well-established classification and regression benchmarks as well as real-world metagenomic datasets. The results show that \ours{} is competitive with, and often improves upon, strong classical and neural FS baselines while selecting fewer features on average across the evaluated benchmarks.
%We share our implementation of \ours{} at \url{https://github.com/wwydmanski/AutoNFS}
\end{abstract}

\section{Introduction}

Feature selection (FS) remains a long-standing challenge in machine learning and data analysis, particularly for high-dimensional tabular datasets, where interpretability and efficiency are crucial~\cite{theng2024feature, dhal2022comprehensive}. In practice, such datasets are often constructed by aggregating all available features or by manually engineering additional ones, which frequently leads to an excessive number of variables, many of which contribute little to downstream tasks. FS addresses this issue by identifying and removing redundant or irrelevant features, thereby improving the interpretability of the model, reducing complexity, and providing clearer insights. Furthermore, training a subsequent prediction model on reduced data helps mitigate model overfitting, reduce variance, and often improve predictive performance.

Existing FS approaches can be broadly categorized into filter~\cite{yu_efcient_nodate,smieja2014asymmetric}, wrapper~\cite{kohavi_wrappers_1997,maldonado_wrapper_2009}, and embedded methods~\cite{regression_shrinkage,zou_regularization_2005}, each with inherent limitations. Filter methods rank features according to statistical relevance but remain independent of the learning model, potentially overlooking complex feature interactions. Wrapper methods iteratively select features using the predictive performance of a model as a criterion, but suffer from high computational costs. 
Embedded methods, such as L1 regularization or attention-based mechanisms, integrate FS within the learning process but may introduce instability or lack fine-grained control over feature importance.
% Existing feature selection approaches can be broadly categorized into filter~\cite{guyon_introduction_nodate,yu_efcient_nodate,smieja2014asymmetric}, wrapper~\cite{kohavi_wrappers_1997,maldonado_wrapper_2009}, and embedded methods~\cite{regression_shrinkage,zou_regularization_2005}, each with inherent limitations. Filter methods may ignore feature interactions and they are agnostic on the predictive model. Wrapper methods tend to be computationally expensive and prone to overfitting. Embedded methods introduce a  bias of predictive model, which makes the selected features narrowed to a certain algorithm. 
The computational cost of most FS algorithms grows rapidly with the number of input dimensions, making them inefficient for large datasets \cite{tan2014towards}. Additionally, the number of selected features is usually treated as a user-defined hyperparameter; an inappropriate choice can lead to suboptimal performance and require multiple retrainings.

To address these limitations, we propose \textbf{\ours{}}, a neural network for efficient and automatic FS. \ours{} is a fully differentiable approach, consisting of two networks trained end-to-end (\Cref{fig:architecture}). The masking network generates a mask that indicates selected features using temperature-controlled Gumbel-Sigmoid sampling \cite{maddison2017concrete, jang2017categorical}, while the target network is a predictive model that evaluates their relevance in a downstream task. Unlike existing methods, where the user must specify the desired number of features, \ours{} automatically determines the minimal subset of features sufficient for the downstream task through a penalty loss component. Moreover, by designing \ours{} as a modern neural network, it avoids repeated retraining across different feature budgets and adds only a lightweight masking module on top of a standard predictor. The total cost still scales with input dimensionality through the predictor, but the \textbf{additional} cost of feature selection is kept small and stable.

We evaluate \ours{} on well-established classification and regression benchmarks with three scenarios of adding corrupted features \cite{cherepanova_performance-driven_nodate}. Our experiments demonstrate that \ours{} consistently outperforms existing techniques while selecting significantly fewer features (\Cref{fig:double_analysis,fig:justavg}). These results are supplemented by the evaluation of \ours{} in real-world metagenomic datasets (\Cref{tab:metagenomic}), an analysis of its computational complexity (\Cref{fig:complexity_estimates,fig:feature_selection_scaling}), and the visualization of its interpretability in the example of the MNIST dataset (\Cref{fig:avg_entropy,fig:entropy_comparison}).

Our contributions can be summarized as follows.
\begin{compactitem}
    \item We propose \ours{}, a novel neural network for end-to-end FS, leveraging Gumbel-Sigmoid relaxation and a regularization term that penalizes the number of selected features.
    
    \item We show that \ours{} automatically identifies a minimal yet sufficient subset of features, achieving a nearly constant computational overhead regardless of the input dimensionality, making it scalable for high-dimensional data.
    
    \item We validate our approach on well-established OpenML-based benchmarks for FS showing its advantage over related methods. In addition, it is examined on real-world metagenomic datasets, highlighting its effectiveness in high-dimensional biological data analysis.
\end{compactitem}

We focus on global feature selection, i.e., learning a single subset shared across the dataset rather than an instance-specific acquisition policy. This design is motivated by settings in which interpretability, fixed inference cost, and reuse of the selected subset are more important than sample-specific adaptation.

\section{Related Work}

In \cite{cheng_comprehensive_2024}, the importance of FS is reviewed broadly, focusing on filter, wrapper, and embedded methods. Similar surveys have emphasized that the basic taxonomy remains relevant, but must now account for the issues of scalability, fairness, and interpretability in modern high-dimensional data analysis \cite{guyon2003introduction,kohavi_wrappers_1997,chandrashekar2014survey,brown2012conditional}. %Due to the page limit, we refer the reader to \Cref{app:rw} for a detailed description of the classical methods.

\paragraph{Classical FS methods}
Filter methods typically rely on statistical criteria such as correlation, mutual information, or significance tests. Classical examples include mRMR \cite{peng2005mrmr}, Relief and its variants \cite{Kononenko1994EstimatingAA,robnik-sikonja_theoretical_2003}, or kernel-based criteria like HSIC Lasso \cite{yamada_high-dimensional_2014}. More recent efforts include measures based on the maximal information coefficient (MIC) to capture non-linear associations \cite{reshef_detecting_2011}. These methods are computationally efficient and easy to interpret, but they ignore feature interactions and are detached from the final predictive model, which often leads to suboptimal subsets.  

Wrapper methods overcome this by iteratively selecting subsets guided by model performance. Classical strategies include sequential forward/backward selection and floating search \cite{Pudil1994FloatingSM}, SVM-RFE for ranking genes \cite{guyon_gene_2002}, and more recent ensemble-based approaches such as Boruta, which compares importance with permuted shadow features \cite{kursa_feature_2010}. Wrappers usually achieve higher accuracy, but their repeated training makes them infeasible for high-dimensional data or large-scale tasks.  

Embedded methods integrate FS directly into the model learning phase. The best known are sparsity-inducing penalties like Lasso \cite{Tibshirani1996RegressionSA}, Elastic Net \cite{zou_regularization_2005}, and Group Lasso \cite{yuan_model_2006,Simon2013ASL}. Tree-based ensembles provide another embedded route: feature importance can be derived from Random Forests \cite{Breiman2001RandomF} or boosting models like XGBoost and CatBoost \cite{Chen:2016:XST:2939672.2939785,prokhorenkova_catboost_2019}. Embedded methods combine efficiency and accuracy, but they are biased toward the structure of the underlying model (linear, tree-based), and may struggle in domains with correlated features. Stability selection was proposed to mitigate these limitations \cite{meinshausen_stability_2009}.

\paragraph{Deep learning FS methods} The rise of deep learning has inspired neural approaches to FS~\cite{ho_adaptive_2021}. Early attempts penalized input weights or used shallow gating networks \cite{li_deep_2016}. Later, continuous relaxations allowed discrete masks to be trained via SGD. \cite{Louizos2017LearningSN} introduced Hard-Concrete gates for $L_0$ regularization; \cite{yamada_feature_2020} proposed Stochastic Gates (STG); and \cite{balin_concrete_2019} designed Concrete Autoencoders that explicitly reconstruct inputs from a subset of features. INVASE \cite{Yoon2018INVASEIV} went further, training an instance-specific selector and predictor in tandem. LassoNet \cite{lemhadri_lassonet_2021} enforced a hierarchical coupling between a linear skip and deep features to guarantee consistency. Attention mechanisms in Transformers have also been used as feature selectors, but their explanatory validity is contested \cite{serrano-smith-2019-attention,jain-wallace-2019-attention,gorishniy_revisiting_2023}. 

Our work builds on this differentiable line. The technical foundation comes from the Gumbel–Softmax trick \cite{jang2017categorical,maddison2017concrete}, which provides low-variance gradients for sampling. This idea has been extended to subset selection through Gumbel-Top-$k$ \cite{kool_stochastic_nodate}, continuous relaxations for sampling without replacement \cite{Xie2019ReparameterizableSS}, and differentiable sorting operators \cite{blondel_fast_nodate}. \cite{strypsteen2024conditionalgumbelsoftmaxconstrainedfeature} proposed Conditional Gumbel–Softmax to incorporate structural constraints into FS, such as sensor topologies. Unlike these, \ours{} addresses unconstrained tabular data and eliminates the need to specify the number of features, letting it emerge from optimization through a cardinality penalty.

Another important line of work studies the acquisition of features \emph{dynamic}, where features have costs and are revealed sequentially. Recent methods query features conditioned on previously observed values \cite{pmlr-v202-covert23a,yasuda_sequential_2023}, or use reinforcement learning to optimize acquisition policies (e.g., EDDI, budgeted classification) \cite{ma_eddi_2019,janisch_classification_2019,trapeznikov_supervised_2013}. These methods are attractive when data acquisition is expensive (medical tests, sensor readings), but they solve a different problem than ours: we focus on learning a single global mask that amortizes selection across all samples, making inference fast and predictable.  

Finally, reliability and fairness in FS have also been addressed. Knockoff-based methods provide false discovery rate control \cite{Barber_2015,Romano_2019}, while stability selection explicitly balances sparsity and robustness \cite{meinshausen_stability_2009}. Greedy and OMP-style selectors have been extended to guarantee approximation bounds and fairness in large-scale problems \cite{pmlr-v206-quinzan23a}. These approaches focus on statistical guarantees, while our method emphasizes efficiency and scalability in neural training.  

% FS is also tightly connected to the broader evolution of tabular learning. Gradient-boosted trees remain strong baselines \cite{Chen:2016:XST:2939672.2939785,prokhorenkova_catboost_2019}, but neural architectures are closing the gap: NODE \cite{popov_neural_2019}, TabTransformer \cite{huang2020tabtransformertabulardatamodeling}, SAINT \cite{DBLP:journals/corr/abs-2106-01342}, and TabNet \cite{arik_tabnet_2020} demonstrate the potential of attention and sparsity in tabular settings. In this context, \ours{} contributes a lightweight, model-agnostic selector: its masks can be plugged into both neural and tree-based backbones. We show that \ours{} consistently reduces dimensionality while maintaining or improving predictive accuracy, and does so with nearly constant computational cost as dimensionality grows.

\section{The proposed model}

In this section, we introduce \ours{}, a neural network approach for automatic selection of features, which are relevant for a given machine learning task. First, we give a brief overview of \ours{}. Next, we describe its main building blocks. Finally, we summarize the training algorithm and the inference phase. %We conclude this section with a discussion of \ours{}.

% \subsection{Gumbel Feature Selection Network}

% \ours{} is a neural network approach to differentiable feature selection. It selects features, which are the most informative for a given classification task. Unlike traditional methods for feature selection that require \marek{separate preprocessing steps (nie rozumiem o co chodzi w tym sformulowaniu, napisz jasniej)} or multiple training iterations, our method integrates feature selection directly into the learning process through a novel application of Gumbel-Sigmoid sampling.

% \subsection{Network Architecture}
% The core of \ours{} consists of two main components:
% \begin{enumerate}
%     \item \textbf{Feature Selection Module}: A learnable embedding vector projects to a feature-sized mask through Gumbel-Sigmoid sampling. This mechanism effectively selects a subset of features while maintaining differentiability.
    
%     \item \textbf{Task Network}: A standard neural network that processes the masked inputs to perform the primary task (classification or regression).
% \end{enumerate}

% The feature selection is achieved through a small yet powerful parameterization. We learn a single embedding vector $e \in \mathbb{R}^{1 \times d_e}$ (where $d_e = 32$ by default) and project it to the feature space dimension using a learnable matrix. This design ensures that the network's parameter count stays nearly constant regardless of input dimensionality, making it particularly efficient for high-dimensional data.

\subsection{Overview of \ours{}}

\ours{} is a neural network that incorporates features selection into a process of learning a predictive model. It retrieves a variable-size subset of attributes that are the most informative for solving a given classification or regression task. 

\begin{figure}[t]
    \centering
    \includegraphics[width=\linewidth]{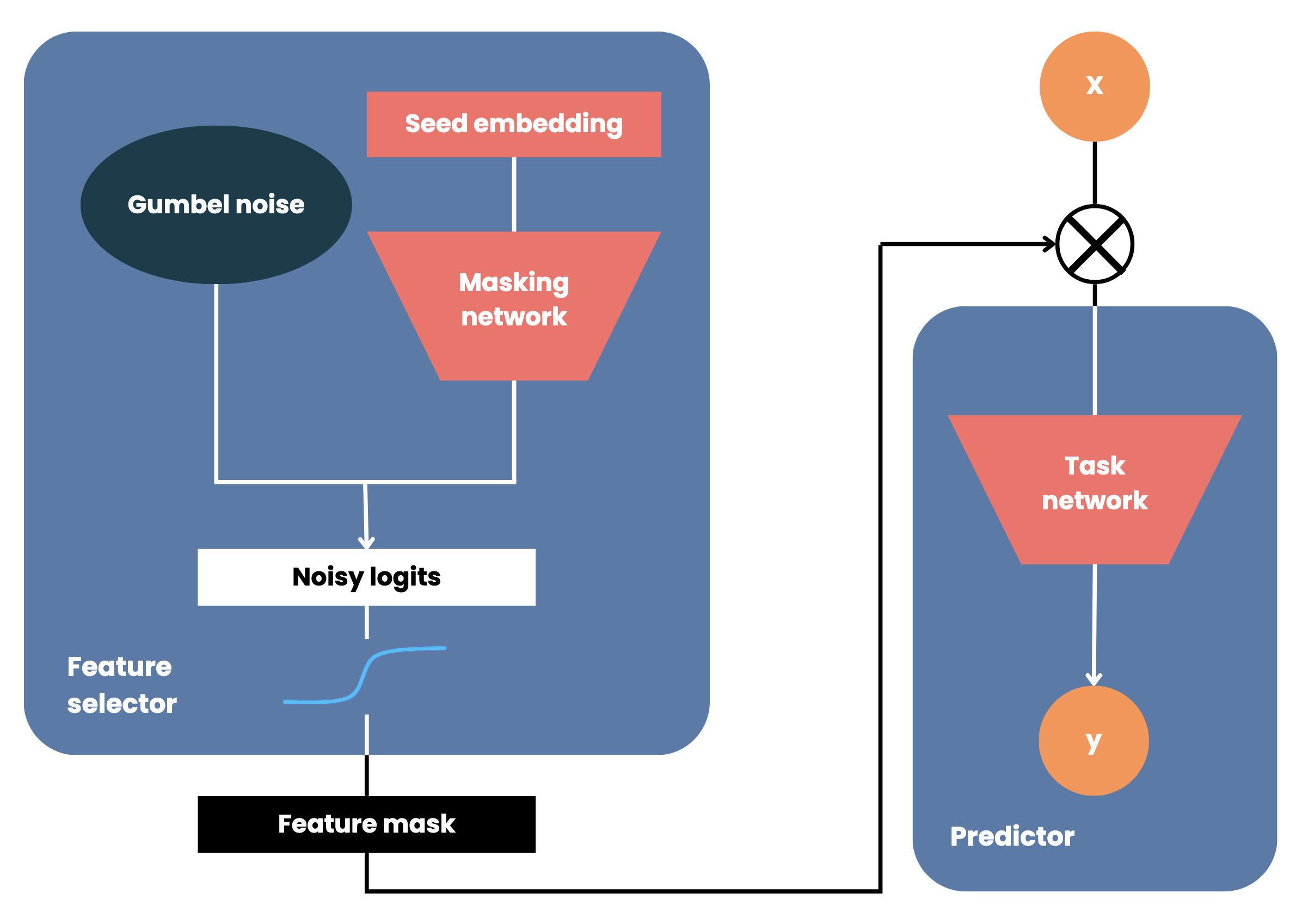}
    \caption{The architecture of \ours{} consists of two parts: masking and task network. The masking network creates a mask representing selected features, while the target network validates these features on a downstream task. The model is trained end-to-end using a differentiable loss and automatically determines the number of output features.}
    \label{fig:architecture}
\end{figure}

The architecture of \ours{} consists of two components: \emph{masking and task networks}, see \Cref{fig:architecture}. While the masking network generates a mask representing selected features, the task network solves the underlying task using the indicated attributes. The loss function of \ours{} combines cross-entropy (for classification) or mean square error (for regression) with the penalty term, which encourages the model to minimize the number of selected features. In consequence, the task network plays the role of a discriminator, which verifies the usefulness of the features chosen for a given task. 

In contrast to traditional methods for FS, which iteratively add or reduce attributes, \ours{} uses a differentiable mechanism to learn a mask based on the Gumbel-Sigmoid relaxation of the discrete distribution~\cite{jang2017categorical,maddison2017concrete}. This design ensures that the computational time remains nearly constant regardless of the input dimensionality, making it particularly efficient for high-dimensional data.

% The following sections describe details behind \ours{}.

\subsection{Masking network}

The masking network $f: \R^{D_e} \to \R^D$ is responsible for generating a mask that indicates features selected for a given dataset $\{(x_i,y_i)\}_{i=1}^N \subset \R^D$. Given a randomly initialized input embedding ${e \in \R^{D_e}}$, the network $f$ outputs $D$-dimensional vector $w = f_{\phi}(e) \in \R^D$, which determines the mask. More precisely, the output vector $w=(w_1,\ldots,w_D)$ is transformed via a sequence of $D$ Gumbel-Sigmoid functions to the (non-binary) mask vector $m=(m_1,\ldots,m_D)$, where ${m_i = GS(w_i; \tau)}$ is given by the Gumbel-Sigmoid function with the temperature parameter $\tau > 0$. Let us recall that the Gumbel-Sigmoid function is given by:
\begin{equation*}
    \text{GS}(w_i; \tau) = \sigma\left(\frac{w_i + g_i}{\tau}\right),
\end{equation*}
where $g_i \sim -\log(-\log(u))$ with $u \sim \text{Uniform}(0,1)$ is the Gumbel noise, $\sigma$ is the sigmoid function, and $\tau > 0$ is the temperature parameter.

For $\tau > 0$, the mask $m=(m_1,\ldots,m_D)$ sampled from the Gumbel-Sigmoid distribution can take a continuous (non-binary) form. As $\tau$ decreases, the mask approaches the binary vector, which represents the final discrete mask. Slow decrease of the temperature $\tau$ allows the model to learn the optimal mask during network training.

\subsection{Task network}

To learn the optimal mask, we need to verify whether it is informative for the underlying task (e.g. classification). To this end, we first apply a mask $m$ to the input example $x$, by element-wise multiplication $x_m = m \odot x$. Next, we feed a task network $g: \R^D \to Y$ with $x_m$ to obtain the final output $g(x_m)$. The relevance of features selected by $m$ is quantified by the cross-entropy or mean-square loss denoted by $\loss_{task}(y; g(x_m))$. Furthermore, to encourage the model to eliminate redundant features, we penalize the model for every added attribute by:
$$
\loss_{select} = \frac{1}{D}\sum_{j=1}^D m_j.
$$
The complete loss function is then given by:
$$
\loss_{total} = \loss_{task} + \lambda \loss_{select},
$$
where $\lambda$ is hyperparameter. We experimentally verified that using a constant value $\lambda=1$ gives satisfactory results across datasets. Thanks to the Gumbel-Sigmoid relaxation of the discrete mask distribution, we can learn the mask during end-to-end differentiable training. 

\subsection{Training process}

%algorithm + loss
Let us summarize the training algorithm described in Algorithm 1. Training starts with a fixed temperature $\tau = \tau_0$ and a randomly initialized embedding $e$. Given an embedding $e$, the masking network $f$ returns a mask vector $m=(m_1,\ldots,m_D)$ using the Gumbel-Sigmoid functions. Each continuous mask vector $m$ sampled from Gumbel-Sigmoid is then applied to a mini-batch $\B$ to construct the reduced vectors $x_m = m \odot x$, for $x \in \B$. This vector goes to the task network $g$, which returns the response for a given task $g(x_m)$. The loss function $\loss_{total}$ is calculated and the gradient is propagated to: (1) embedding vector $e$, (2) weights of $f$ and $g$. In particular, by learning the embedding vector $e$ and the parameters of $f$, we optimize the mask vector.

\begin{algorithm}[ht]
\caption{\ours{} training procedure for classification}
\begin{algorithmic}[1]
\footnotesize
\State \textbf{Input:} Dataset $\mathcal{D} = \{(\mathbf{x}_i, \mathbf{y}_i)\}_{i=1}^N$, batch size $B$, initial temperature $\tau_0 = 2.0$, decay rate $\alpha = 0.997$, total epochs $E$, FS balance parameter $\lambda$
\State \textbf{Initialize:} Embedding vector $\mathbf{e} \in \mathbb{R}^{d_e}$, masking network $f_{\phi}$, task network $g_{\theta}$
\State $\tau \leftarrow \tau_0$
\For{epoch $= 1$ to $E$}
    \For{each mini-batch $\mathbf{\B} = \{(\mathbf{x}_i, \mathbf{y}_i)\}_{i=1}^B \subset \mathcal{D}$}
        \State $\mathbf{w} \leftarrow f_\phi(\mathbf{e})$ \Comment{Compute logits for feature mask}
        \State $\mathbf{g} \leftarrow -\log(-\log(\mathbf{u}))$, where $\mathbf{u} \sim \text{Unif}(0,1)$ \\ \Comment{Sample Gumbel noise}
        \State $\mathbf{m} \leftarrow \sigma\left((\mathbf{w} + \mathbf{g})/\tau\right)$\\ \Comment{Generate mask via Gumbel-Sigmoid}
        \State $\mathbf{X} \leftarrow \{\mathbf{x}_i\}_{i=1}^B$
        \State $\mathbf{X}_{\text{masked}} \leftarrow \mathbf{X} \odot \mathbf{m}$ \Comment{Mask input features}
        
        \State $\hat{\mathbf{Y}} \leftarrow g_{\mathbf{\theta}}(\mathbf{X}_{\text{masked}})$ \\ \Comment{Forward pass through task network}
        \\
        
        \State $\mathcal{L}_{\text{task}} \leftarrow -\sum_{i=1}^B \sum_{c=1}^C y_{i,c} \log(\hat{y}_{i,c})$
        % \frac{1}{B} \sum_{i=1}^B (y_i - \hat{y}_i)^2 & \text{for regression}
        % \end{cases}$
        
        \State $\mathcal{L}_{\text{select}} \leftarrow \frac{1}{D}\sum_{j=1}^{D} m_j$
        
        \State $\mathcal{L}_{\text{total}} \leftarrow \mathcal{L}_{\text{task}} + \lambda \cdot \mathcal{L}_{\text{select}}$
        \\
        
        \State $\mathbf{e} \leftarrow \mathbf{e} - \eta_1 \nabla_{\mathbf{e}} \mathcal{L}_{\text{total}}$ \Comment{Update embedding}
        \State $\phi \leftarrow \phi - \eta_1 \nabla_{\phi} \mathcal{L}_{\text{total}}$ \Comment{Update masking network}
        \State $\theta \leftarrow \theta - \eta_2 \nabla_{\theta} \mathcal{L}_{\text{total}}$ \Comment{Update task network}

    \EndFor
    \State $\tau \leftarrow \tau \cdot \alpha$ \Comment{Anneal temperature}
\EndFor
\end{algorithmic}
\end{algorithm}

% Making use of non-zero temperature $\tau$, this approach allows the network to initially explore the feature space broadly and gradually commit to specific features as $\tau$ decreases. The feature selection regularization term $\loss_{select}$ can be configured in two modes: either penalizing deviation from a target number of features, or directly encouraging sparsity by minimizing the number of active features. 

A critical aspect of our algorithm is the temperature annealing schedule. We begin with a high temperature ($\tau=2.0$), which produces soft masks that allow gradient flow to all features. As training progresses, the temperature decays exponentially (typically with $\alpha=0.997$), causing the masks to become increasingly binary.
This gradual transition serves multiple purposes:

\begin{compactitem}
\item It allows the network to initially explore the full feature space.
\item It enables progressive commitment to more discrete FS decisions.
\item It leads to convergence on a nearly binary FS mask at the end of training.
\end{compactitem}

The annealing process effectively functions as a curriculum, starting with easier optimization (continuous selection) and progressively transitioning to harder optimization (discrete selection). This process is related to exploration-exploitation trade-off, which parallels fundamental concepts in reinforcement learning.

\subsection{Feature Importance Quantification}
After training, we quantify the importance of each feature by directly applying the learned selection mechanism with hard Gumbel-Sigmoid activation:
\begin{compactenum}
    \item Calculating the feature logits of the trained embedding: $\mathbf{w} = \mathbf{f}_{\phi}(\mathbf{e})$.
    \item Applying a hard threshold, that is, if $\sigma(w_i) > 0.5$, then $m_i=1$, else $m_i = 0$.
    \item Interpreting the resulting binary vector ${\mathbf{m}=(m_1,\ldots,m_D)}$ as the mask for feature selection.
\end{compactenum}

This process produces a deterministic FS that clearly identifies relevant features for the task. Since our FS mechanism is parameterized by a single embedding vector that is independent of specific input examples, the selected features remain constant throughout the dataset. The resulting binary mask can be directly used to filter features, or features can be ranked by their logit values when a specific top-$k$ selection is desired. Importantly, since the selection mechanism was jointly optimized with the task objective, the selected features capture both individual importance and interactive effects relevant to the specific task.

\section{Experiments}

\begin{figure*}
  \centering
  \includegraphics[width=0.8\linewidth]{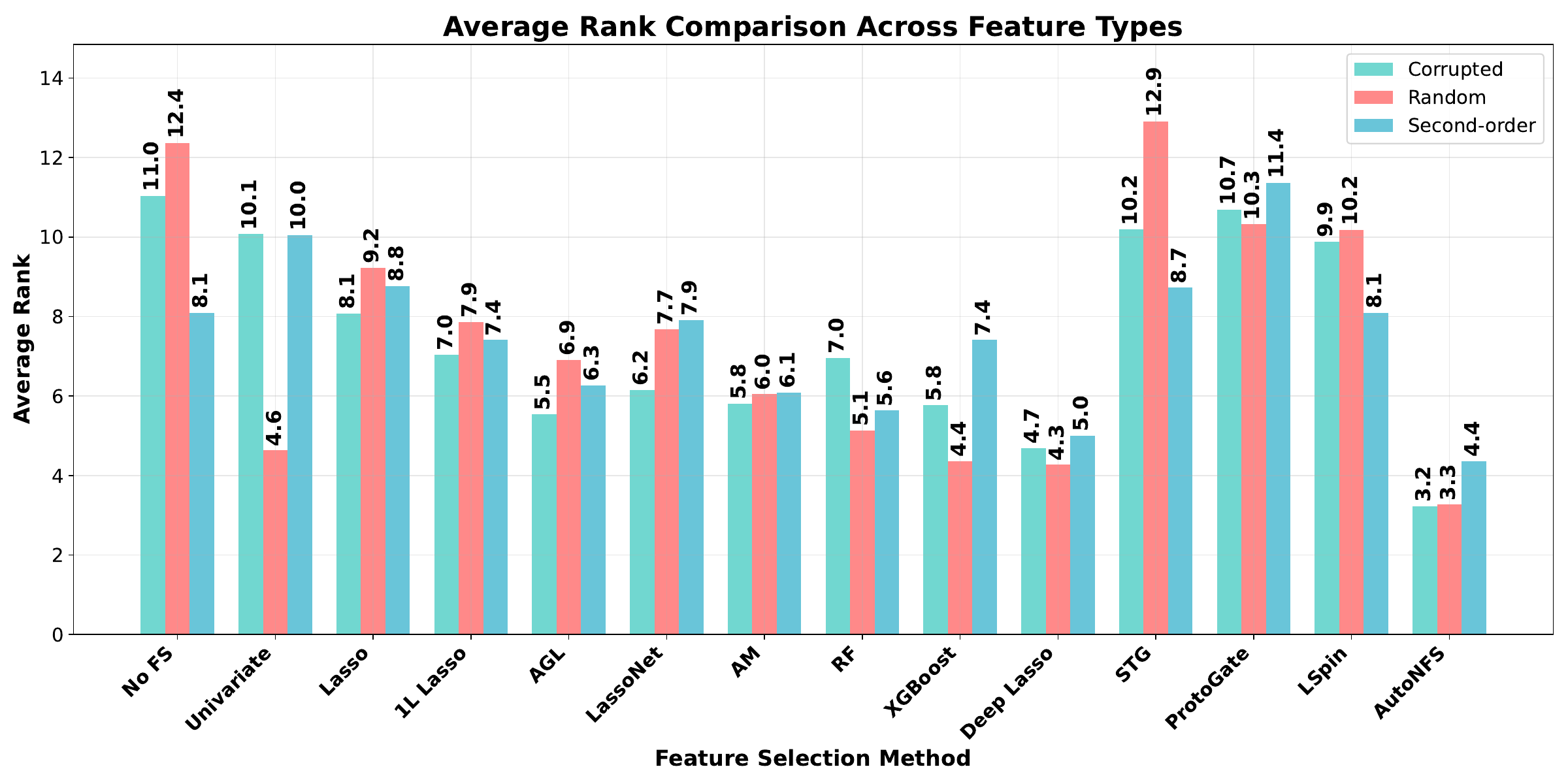}
  \caption{Average ranking of FS methods for three types of features corruption shows that \ours{} consistently outperforms all competitive methods.}
  \label{fig:justavg}
\end{figure*}

\begin{table}[!htb]
\centering
\scriptsize
% \tiny
\setlength{\tabcolsep}{3pt}
\caption{The number of attributes selected by \ours{} under three considered scenarios. It is evident that \ours{} not only eliminate auxiliary noisy features but also drastically reduces the number of the original attributes.}
\begin{tabular}{lccc}
\toprule
\textbf{Dataset} & \textbf{Random} & \textbf{Corrupted} & \textbf{Second-order} \\
& \textbf{Features} & \textbf{Features} & \textbf{Features} \\
\midrule
AL (aloi)  & 65 & 65 & 69 \\
CH (california) & 5 & 5 & 3 \\
EY (eye) & 8 & 11 & 12 \\
GE (gesture) & 11 & 16 & 22 \\
HE (helena)  & 15 & 14 & 16 \\
HI (higgs\_small) & 14 & 14 & 14 \\
HO (house)  & 10 & 10 & 9 \\
JA (jannis)  & 17 & 16 & 18 \\
MI (microsoft)  & 47 & 61 & 42 \\
OT (otto) & 78 & 67 & 76 \\
YE (year) & 69 & 28 & 29 \\
\bottomrule
\end{tabular}
\label{tab:statsMain}
\end{table}

% \begin{figure}
%     \centering
%     \includegraphics[width=\linewidth]{average_rank_comparison.pdf}
%     \caption{Caption}
%     \label{fig:rank_comparison}
% \end{figure}

% \begin{figure}
%     \centering
%     \includegraphics[width=\linewidth]{performance_radar_chart.pdf}
%     \caption{Caption}
%     \label{fig:radar}
% \end{figure}

% % Moved radar chart to first page as first graph.

\begin{table}[t]
    \caption{Performance on metagenomic data reduced with \ours{}. Although \ours{} heavily reduces data dimensionality, it does not lead to the deterioration of the results on average. Each dataset's name is derived from the first author's surname and the year of publication.}
\scriptsize
\setlength{\tabcolsep}{1pt}
    \centering
    \begin{tabular}{l|cc|cc|c}
\toprule
  & MLP on & MLP on & RF on & RF on  & Reduction  \\
dataset & full data & \ours{} & full data & \ours{} & rate \\
\midrule
NielsenHB\_2014 	& 61.3				& \textbf{64.3} &	\textbf{71.1} & 63.4		& 91\% \\
WirbelJ\_2018 		& 55.8				& \textbf{57.1} &	77.6 & \textbf{82.1}		& 94\% \\
% KeohaneDM\_2020 	& \textbf{46.9} 	& 34.4 		 &	46.9 & \textbf{53.1}		& 540 & 37 \\
JieZ\_2017 			& \textbf{69.3} 	& 61.2 		 &	76.2 & \textbf{77.0}		& 80\%\\
FengQ\_2015 		& \textbf{66.2} 	& 60.7 		 &	83.3 & \textbf{88.9}		& 95\%\\
ThomasAM\_2019c 	& 58.2 			& \textbf{66.4} &	62.7 & \textbf{76.4}		& 92\% \\
% LiJ\_2017 			& 34.1 			& \textbf{51.1} &	\textbf{56.1} & 43.2		& 651 & 43 \\
ZellerG\_2014 		& 61.4 			& 61.4 		 &	65.2 & \textbf{87.1}		& 96\% \\
% LifeLinesDeep\_2016 & 51.3 			& \textbf{54.6} &	\textbf{50.0} & \textbf{50.0}		& 526 & 79 \\
ThomasAM\_2018b 	& \textbf{68.6} 	& 61.4 		 &	\textbf{58.6} & \textbf{58.6}		& 95\% \\
HanniganGD\_2017 	& 46.7 			& \textbf{63.3} &	\textbf{81.7} & 53.3		& 95\% \\
YachidaS\_2019 		& 47.1 			& \textbf{57.0} &	\textbf{63.6} & 60.8		& 81\% \\
ZhuF\_2020 			& \textbf{65.7} 	& 55.9			 &	\textbf{76.8} & 73.9		& 95\%\\
ThomasAM\_2018a 	& \textbf{73.3} 	& 56.7			 &	81.7 & \textbf{91.7}		& 91\% \\
% LiJ\_2014 			& 45.4 			& \textbf{49.0} &	50.0 & \textbf{50.8}		& 503 & 46 \\
LeChatelierE\_2013 	& \textbf{55.1} 	& 52.1 		 &	54.9 & \textbf{62.0}		& 92\% \\
QinN\_2014 			& 74.6 			& \textbf{81.5} &	83.3 & \textbf{85.5}		& 94\% \\
QinJ\_2012 			& 55.1 			& \textbf{56.1} &	61.6 & \textbf{62.2}		& 86\% \\
NagySzakalD\_2017 	& 52.1 			& \textbf{58.3} &	91.7 & \textbf{95.8}		& 96\% \\
% YuJ\_2015 			& \textbf{65.3} 	& 41.7 		 &	\textbf{67.4} & 64.6		& 606 & 34 \\
GuptaA\_2019 		& 81.2 			& \textbf{93.8} &	87.5 & \textbf{93.8}		& 97\% \\
VogtmannE\_2016 	& 66.7 			& \textbf{68.1} &	\textbf{69.4} & \textbf{69.4}		& 90\% \\
% AsnicarF\_2021 		& 50.3 			& \textbf{52.8} &	\textbf{50.0} & \textbf{50.0}		& 537 & 90 \\
RubelMA\_2020 		& 60.7 			& \textbf{71.7} &	77.5 & \textbf{79.6}		& 95\% \\
\midrule
average & 62.17 & 63.72 & 73.57 & 75.63 & 92\% \\
\bottomrule
\end{tabular}
\label{tab:metagenomic}

\end{table}

To evaluate the effectiveness of \ours{}, we conducted extensive experiments across multiple datasets (standard OpenML data and high-dimensional metagenomic datasets) and compared our approach with state-of-the-art FS methods, \Cref{sec:bench,sub:metagenomics}. We verify the performance of the model and inspect the importance of selected attributes. Furthermore, we analyze the computational efficiency of our method compared to existing approaches, \Cref{sec:time} and the influence of the parameter $\lambda$ on the behavior of the algorithm, \Cref{app:lambda}. We also provide further insight into the interpretability of the selected features in the example of MNIST, which can be found in \Cref{app:mnist}.

% \marek{powidziec ze robimy jeszcze analizy (jakie)}

% \marek{dopisac tu ze czesc wyników jest w appendix (ze wzgledu na brak przestrzeni w pracy)}

\subsection{Feature Selection Benchmark} \label{sec:bench}

We follow a recent benchmark introduced in \cite{cherepanova_performance-driven_nodate}. The reported results were achieved by extending their codebase with \ours{}.

Following the benchmark protocol, the competing selectors are evaluated at a fixed feature budget equal to the dimensionality of the original, uncorrupted representation. In contrast, \ours{} is not given this budget and instead learns the effective cardinality through the sparsity term.

\paragraph{Datasets}

The benchmark consists of three scenarios applied to 11 datasets, see LHS of \Cref{tab:stats}. In each scenario, a given dataset is corrupted by adding auxiliary features: (1) fully random features, (2) original features corrupted with Gaussian noise, and (3) a set of second-order features created by multiplying randomly selected features from the original dataset. We analyze the case in which 50\% of the features were artificially created in each dataset. By applying FS algorithms, we aim to eliminate redundant features without compromising the predictive power of the representation.

\begin{figure*}[t]
    \centering
    \begin{subfigure}[t]{0.48\linewidth}
        \centering
        \includegraphics[width=\linewidth]{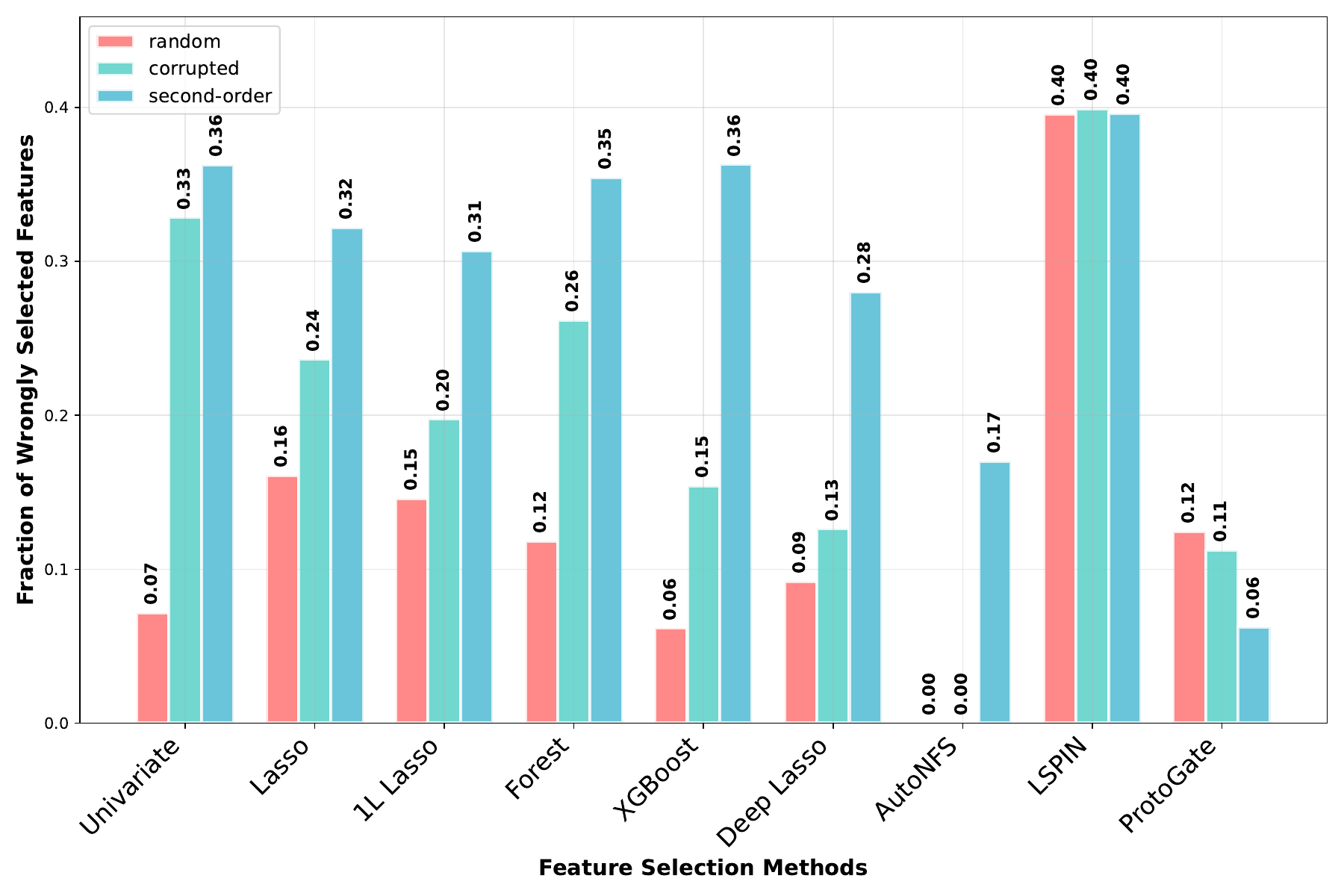}
        \caption{Feature misselection errors. In 2 out of 3 corruption scenarios, \ours{} selects only features from the original ones, presenting the best performance in all cases.}
        \label{fig:wrongly_selected}
    \end{subfigure}
    \hfill
    \begin{subfigure}[t]{0.48\linewidth}
        \centering
        \includegraphics[width=\linewidth]{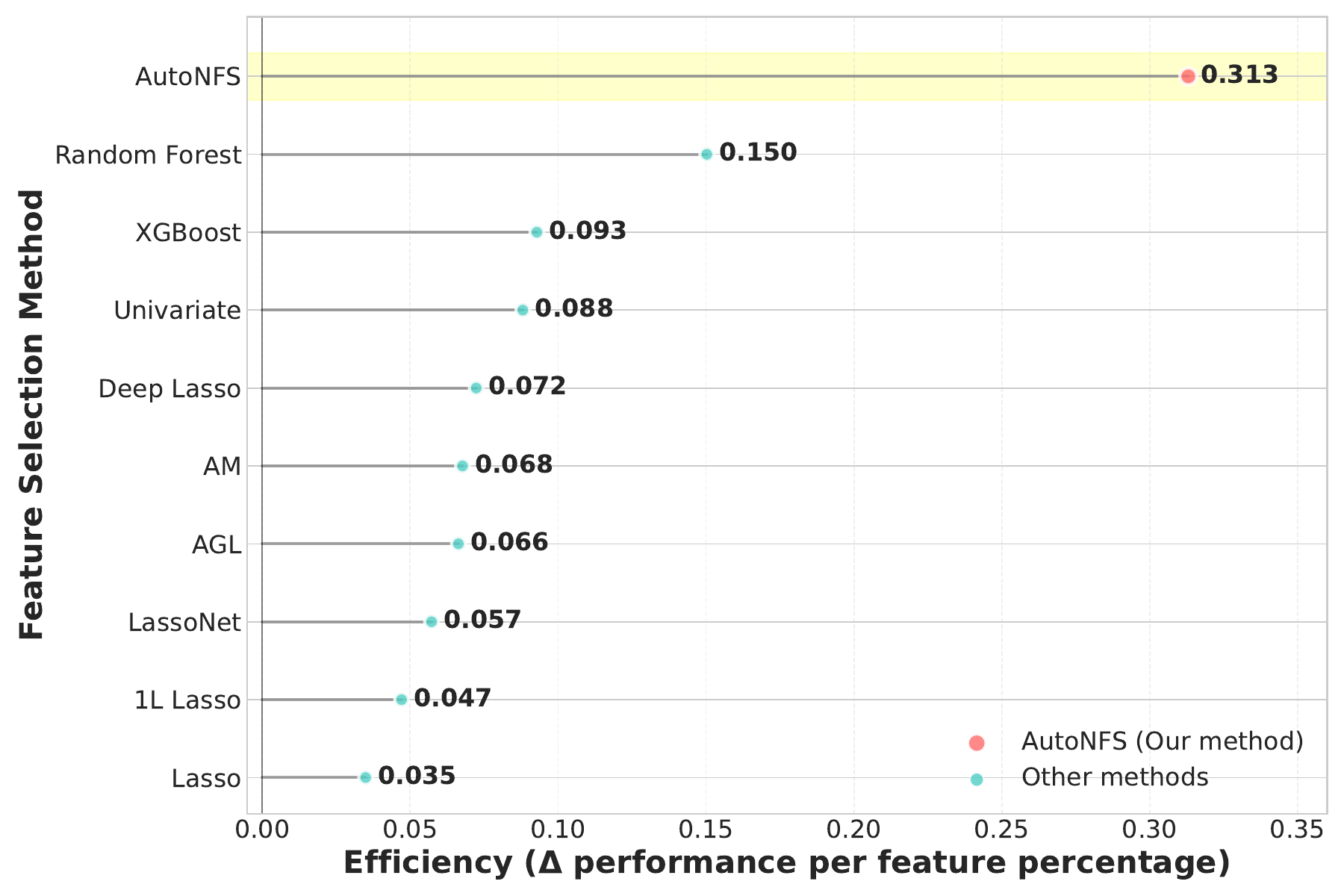}
        \caption{Average predictive power of the selected variables in the case of random feature corruption shows that \ours{} selects the most essential features -- the average performance of downstream model would decrease by 0.313 if any of features selected by \ours{} were eliminated.}
        \label{fig:lollipop_random}
    \end{subfigure}
    \caption{Analysis of features selected by examined methods.
    %Feature misselection for different types of artificial noise. (Right) Predictive power of variables selected in the random features scenario.
    }
    \label{fig:double_analysis}
\end{figure*}

\paragraph{Baseline Methods}

We compared \ours{} with 12 established FS methods:
\begin{compactitem}
    \item No Feature Selection (No FS),
    \item Univariate Selection: Statistical tests for feature ranking (F-statistics),
    \item Lasso: L1-regularized linear models,
    \item 1L Lasso: Single-layer neural network with L1 regularization,
    \item AGL: Adaptive Group Lasso~\cite{ho_adaptive_2021},
    \item LassoNet: Neural network with hierarchical sparsity~\cite{lemhadri_lassonet_2021},
    \item AM: Attention Maps for feature importance~\cite{gorishniy_revisiting_2023},
    \item RF: Random Forest importance,
    \item XGBoost: Gradient boosting importance~\cite{Chen:2016:XST:2939672.2939785},
    \item Deep Lasso: Deep neural network with L1 regularization~\cite{cherepanova_performance-driven_nodate},
    \item STG: Feature Selection using Stochastic Gates \cite{yamada2020featureselectionusingstochastic},
    \item ProtoGate: Prototype-based Neural Networks with Global-to-local Feature Selection for Tabular Biomedical Data \cite{pmlr-v235-jiang24c},
    \item LSpin: Locally Sparse Neural Networks for Tabular Biomedical Data \cite{yang2022locallysparseneuralnetworks}.
\end{compactitem}

The results of each method are evaluated by applying MLP classifier or regressor and reporting performance metrics specific to the task (accuracy for classification, negative mean squared error for regression). We also report the mean rank across datasets to provide an overall performance assessment.

All methods were optimized using 50 steps of Optuna with maximum runtime of 48 hours on a single GH200 GPU.

\paragraph{Model Architecture and Hyperparameters}

\ours{} consists of a 32-dimensional learnable embedding that projects to feature-specific selection logits through a linear layer (32 $\rightarrow$ input\_size), followed by a 3-layer task network with architecture input\_size $\rightarrow$ 32 $\rightarrow$ 32 $\rightarrow$ output\_size using ReLU activations. Hyperparameters were optimized using Optuna~\cite{akiba2019optunanextgenerationhyperparameteroptimization} across epochs $\in$ \{10, 20, 50, 100, 200, 300, 400\}, temperature decay $\in$ \{0.995, 0.997, 0.999\}, and batch sizes $\in$ \{32, 64, 128\}.%, with target features mode selected from \{"raw", "target"\}. 

We use the Adam~\cite{kingma2017adammethodstochasticoptimization} optimizer with separate learning rates: 4e-3 for the FS component and 3e-4 for the task network. The Gumbel-Sigmoid temperature starts at 2.0 and decays per epoch, while the FS balance parameter $\lambda$ is set to 1.0. This design ensures nearly constant computational overhead regardless of input dimensionality while maintaining effective FS capabilities.

\begin{figure*}[t]
    \centering
    \begin{subfigure}[t]{0.48\linewidth}
        \centering
        \includegraphics[width=\linewidth]{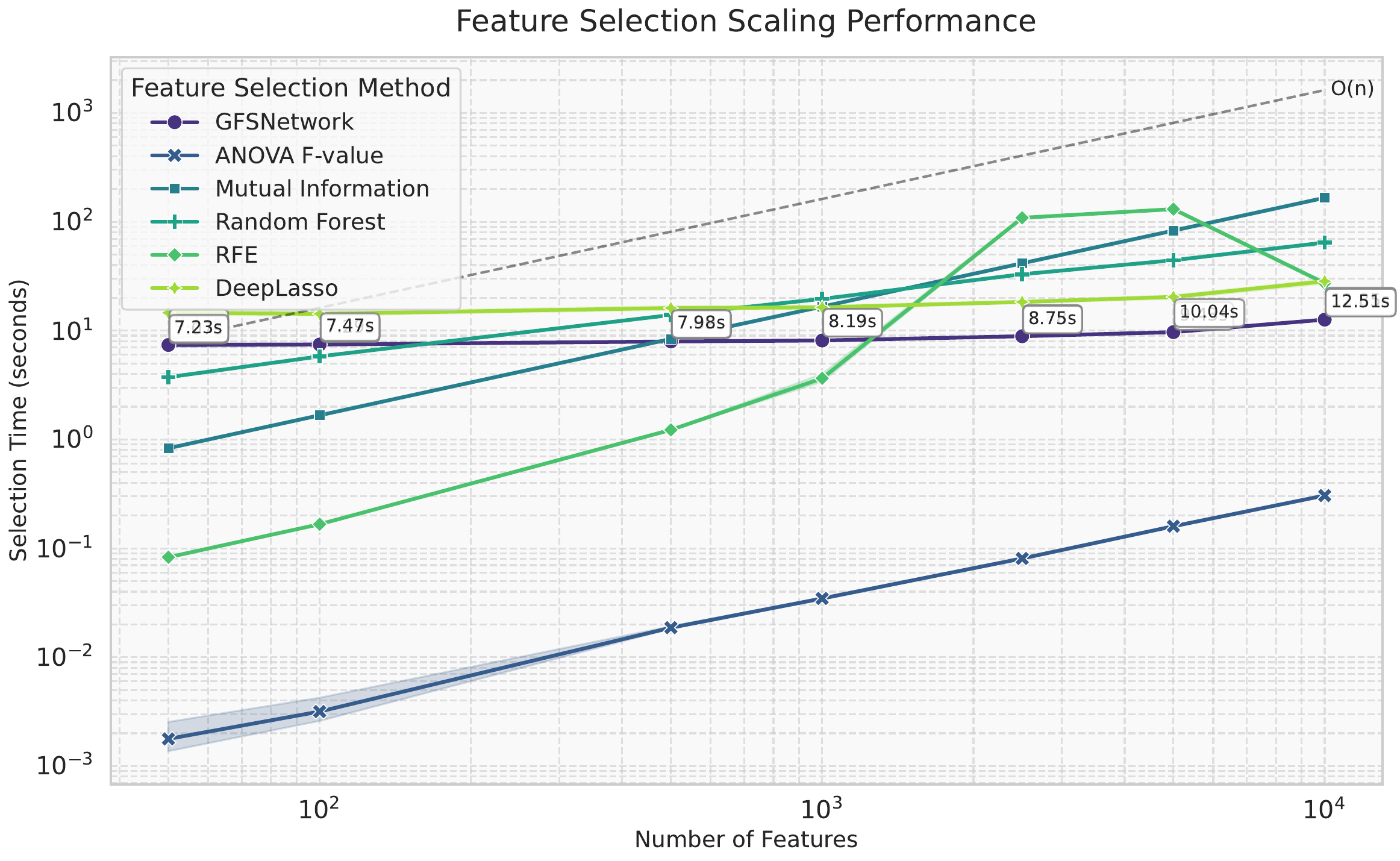}
    \caption{The time requirements of \ours{} remains almost constant with increasing number of features.}
    \label{fig:feature_selection_scaling}
    \end{subfigure}
    \hfill
    \begin{subfigure}[t]{0.48\linewidth}
        \centering
       \includegraphics[width=\linewidth]{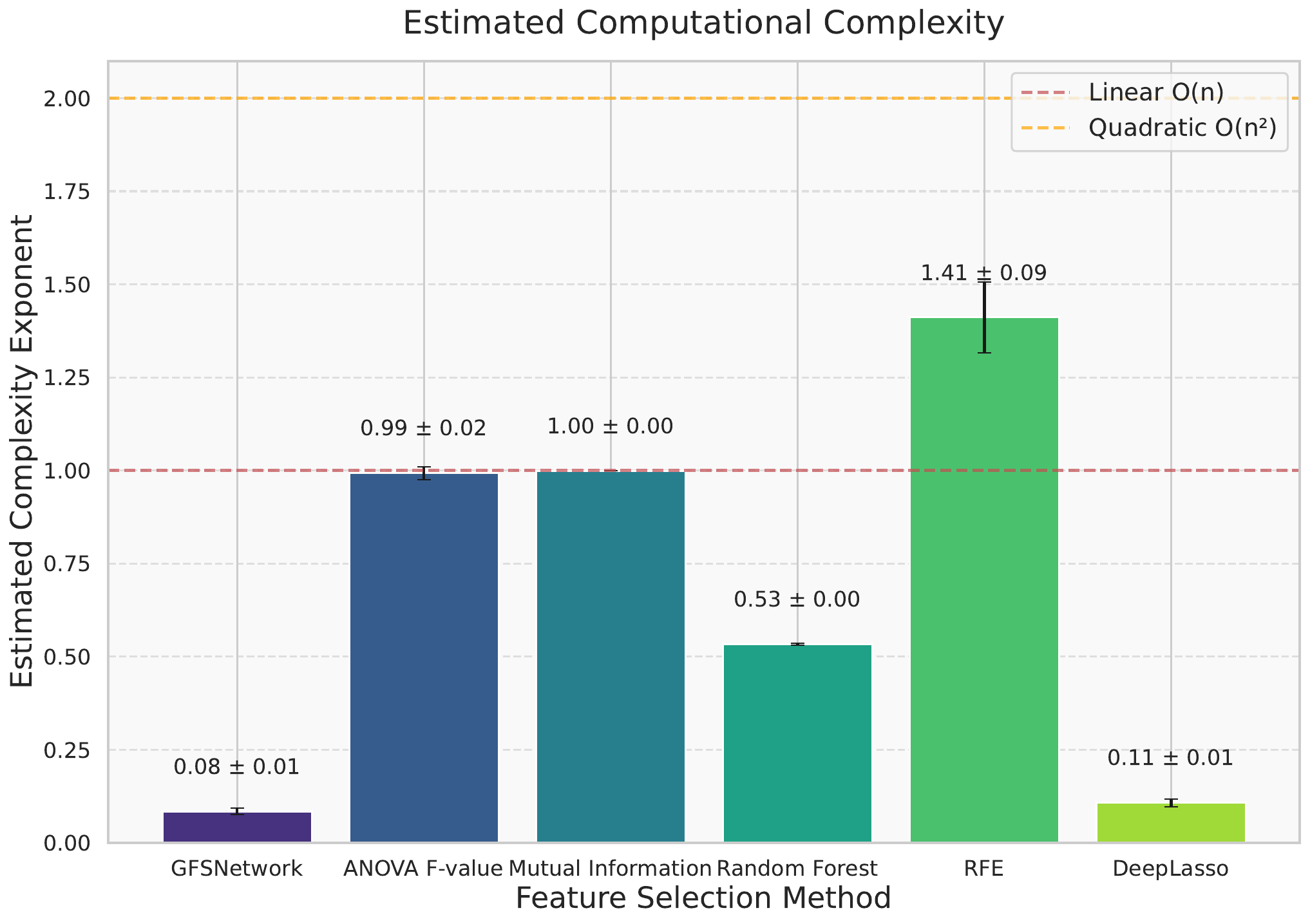}
    \caption{Comparison of estimated complexity exponent for different FS methods as dimensionality changes.}
    \label{fig:complexity_estimates}
    \end{subfigure}
    \caption{Estimation of time complexity.}
    % \label{fig:double_analysis}
\end{figure*}

\paragraph{Predictive performance}

% \marek{brakuje informacji ze kazda z metod wybiera tyle cech ile bylo oryginalnie w datasecie a nasza sama decycuje. Podkreslic potem ze ilosc wybranych przez nasza metode cech jest zawsze mniejsza niz u innych metod co sprawia ze nasza metoda jest istotnie lepsza}

The ranking summary of the results presented in \Cref{fig:justavg} shows an impressive performance of \ours{} in each scenario. While the highest advantage of \ours{} is observed for the case of features corrupted by Gaussian noise (average rank 2.1), in the remaining two scenarios (random and second-order features) \ours{} still achieves the best ranks, beating the next competitors by 1.0 and 0.6 ranking points, respectively. It is important to note that all baseline methods select the same number of features as were in the initial representation (before corruption), whereas our method automatically chooses a much smaller subset of the most relevant features, see \Cref{tab:statsMain}. As a result, \ours{} consistently achieves competitive or superior performance while using significantly fewer features, highlighting its practical advantage. Detailed results presented in \Cref{tab:random_features,tab:corrupted_features,tab:secondorder_features} show that our algorithm obtains the highest or joint-highest scores on most datasets, demonstrating consistent and strong performance.

\paragraph{Analysis of selected features} 

In addition to predictive performance on downstream tasks, we analyze how the selected attributes match the original features (before adding auxiliary features). \Cref{fig:wrongly_selected} shows that \ours{} achieves zero misselection errors for random and corrupted features and maintains low error rates of 0.17 for second-order features. It is important to note that the selection of features outside the original attributes in the latter case is acceptable since additional features were created by multiplying the original features. In consequence, these extra features may sometimes carry even more information than the individual original attributes. The application of the representation created by the baseline methods resulted in significantly higher misselection errors.  

% the fraction of features that were incorrectly selected for different feature selection methods in the three experimental scenarios. \ours{} achieves zero misselection errors for random and corrupted features and maintains low error rates of 0.17 for second-order features, which were impossible for all competitors. It is important to note that selecting features outside original attributes in the latter case is acceptable since additional features were created by multiplying original features. In consequence, these extra features may sometimes carry even more information than individual original attributes.

\Cref{fig:lollipop_random} presents the average predictive power of the individual features. More precisely, we measure how much predictive performance decreases when we remove one of the selected features. As can be seen, the average decrease for \ours{} is equal to 0.313, which means that the returned set cannot be further reduced without affecting predictive performance. This demonstrates the superior precision of \ours{} in identifying relevant features while automatically determining the optimal number to select. 

In general, these findings confirm that \ours{} is broadly applicable to a wide range of machine learning tasks, including both classification and regression, while offering strong and reliable performance in various feature noise scenarios.

\subsection{Metagenomic Dataset Analysis}
\label{sub:metagenomics}

\paragraph{Evaluation} To evaluate \ours{}'s effectiveness in real-world high-dimensional biological data, we applied it to 18 metagenomic datasets obtained from Curated Metagenomics Data~\cite{curated_metagenomics}. These datasets represent a particularly challenging domain with high feature dimensionality (308-718 features) and complex biological interactions. In this experiment, we additionally verify how the constructed representation is useful for two types of downstream classifiers: MLP and Random Forest (RF).

% Both versions, before and after feature selection, use MLP as a downstream classifier.

% The results presented in \Cref{tab:metagenomic} demonstrate that on average \ours{} maintains the performance on downstream tasks (improvement of 0.7 pp in accuracy) while drastically reducing feature dimensionality (\ours{} selected only 7.7\% of the original features). More detailed inspection reveals that in 16 out of 24 datasets, our method improved or maintained performance compared to using all features. 

% \Cref{fig:evolution} illustrates the process of feature selection. Observe that \ours{} deeply explores the space of all features in the training phase and selects the final set of features at the end of the training. 

The results presented in \Cref{tab:metagenomic} demonstrate that, on average, \ours{} improved predictive performance on downstream tasks while drastically reducing feature dimensionality. In the case of MLP, \ours{} achieved 1.6 improvements in pp accuracy, while for RF the improvement increased to 2.1 pp, reducing 92\% of the original features. This means that the high predictive performance of the representation generated by \ours{} is independent of a downstream classifier.

\subsection{Computational Complexity Estimation} \label{sec:time}

% The estimated computational complexity reveal striking differences between FS methods, see \Cref{fig:feature_selection_scaling}. Denoting time complexity as an exponential function of the number of features $t \approx D^\alpha$, our empirical analysis shows that \ours{} demonstrates near-constant time scaling ($\alpha \approx 0.08$).  %significantly outperforming all competing approaches as dimensionality increases. 
% Conventional FS methods, such as the ANOVA F value and Mutual Information, exhibit linear scaling ($\alpha \approx 1.0$), while Random Forest FS shows sublinear behavior ($\alpha \approx 0.53$). In contrast, Recursive Feature Elimination with Linear SVC demonstrates superlinear scaling ($\alpha \approx 1.41$), causing its performance to degrade more rapidly with increasing feature dimensions. 

% The confidence intervals over 5 runs (\Cref{fig:complexity_estimates}) indicate that these estimates are statistically robust across the dimensionalities tested. This assessment provides compelling evidence for the exceptional efficiency advantage of \ours{} in high-dimensional FS tasks, with its nearly constant-time behavior representing a significant algorithmic advancement over conventional methods.
Figure~\ref{fig:feature_selection_scaling} reports an empirical wall-clock scaling analysis rather than a formal complexity theorem. Among the compared methods, \ours{} exhibits the shallowest growth with the number of features in the tested regime. We attribute this behavior to the fact that \ours{} learns a mask in a single end-to-end optimization run, without repeated retraining across candidate feature budgets. We therefore interpret these results as evidence of favorable practical scaling in our setup, rather than as a claim of dimension-independent theoretical complexity.

For reference, classical filter methods such as ANOVA F-value and mutual information exhibit approximately linear scaling with the number of features, while wrapper-style approaches tend to scale less favorably. The confidence intervals over 5 runs (\Cref{fig:complexity_estimates}) indicate that the observed trends are consistent across the tested dimensionalities.

\subsection{Influence of the balance parameter} \label{app:lambda}

% \marek{TODO - opis bo na razie nie ma wykresu....}

\begin{figure}
    \centering
    \includegraphics[width=0.9\linewidth]{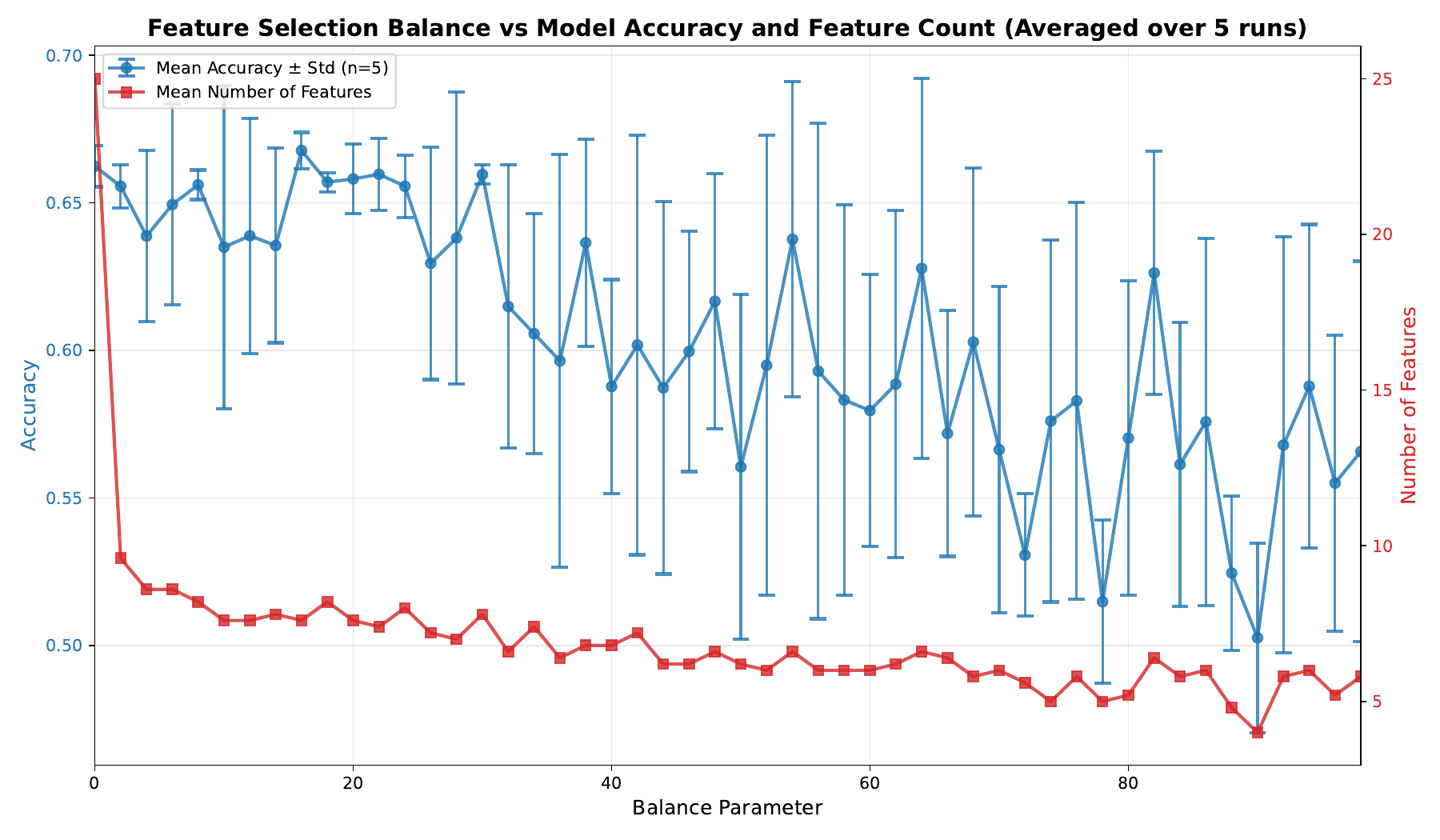}
    \caption{Effect of the balance parameter $\lambda$ on predictive accuracy (blue, left axis) and the number of selected features (red, right axis). When $\lambda=0$, \ours{} prioritizes task performance and selects a large number of features. As $\lambda$ increases, the sparsity penalty reduces the number of features while preserving accuracy up to a point. Very high values of $\lambda$ cause over-sparsification, where too few features are selected, leading to performance degradation. Results are averaged over 5 runs with standard deviations shown as error bars.}

    \label{fig:placeholder}
\end{figure}

The balance parameter $\lambda$ in \ours{} controls the trade-off between task performance and feature sparsity through the regularization term $\lambda \mathcal{L}_{select}$ in the total loss function, see \Cref{fig:placeholder}. When $\lambda = 0$, the model prioritizes the performance of tasks without penalizing the use of features, typically selecting a larger number of features. As $\lambda$ increases, the sparsity penalty becomes more influential, forcing the model to select fewer features while trying to maintain predictive accuracy. However, excessively high values of $\lambda$ lead to over-sparsification, where the model selects too few features to adequately capture the underlying patterns, resulting in performance degradation. This analysis demonstrates the importance of proper tuning of $\lambda$ and highlights how \ours{} can automatically navigate the accuracy-sparsity trade-off to identify optimal feature subsets in different datasets.

% \begin{figure}[t]
%     \centering
%     \includegraphics[width=\textwidth]{mnist_selected_features_comparison.pdf}
%     \caption{Analysis of sample features (top-left) from MNIST dataset shows that entropy of selected features (F1-F3) is much higher than their non-selected counterparts (F4, F5). It confirms that \ours{} selects the most discriminative features.}
%     \label{fig:mnist}
% \end{figure}
% \clearpage
% \section{Discussion}

% \section{Conclusion}

% We presented \ours{}, a novel neural architecture for differentiable feature selection using temperature-controlled Gumbel-Sigmoid sampling. Our experiments demonstrate that \ours{} offers a compelling approach to feature selection, particularly for classification tasks with high-dimensional data. The method's near-constant computational scaling ($\alpha \approx 0.08$) represents a significant advantage over traditional approaches, making it particularly valuable for datasets with thousands of features -- such as omics datasets. The end-to-end learning approach allows \ours{} to optimize feature selection specifically for the task at hand, while automatically determining the appropriate number of features to select. As demonstrated in our MNIST visualization experiments, the selected features align with human intuition, providing interpretability that is valuable in domains like healthcare and biology.

\begin{figure}[t]
    \centering
    \includegraphics[width=\linewidth]{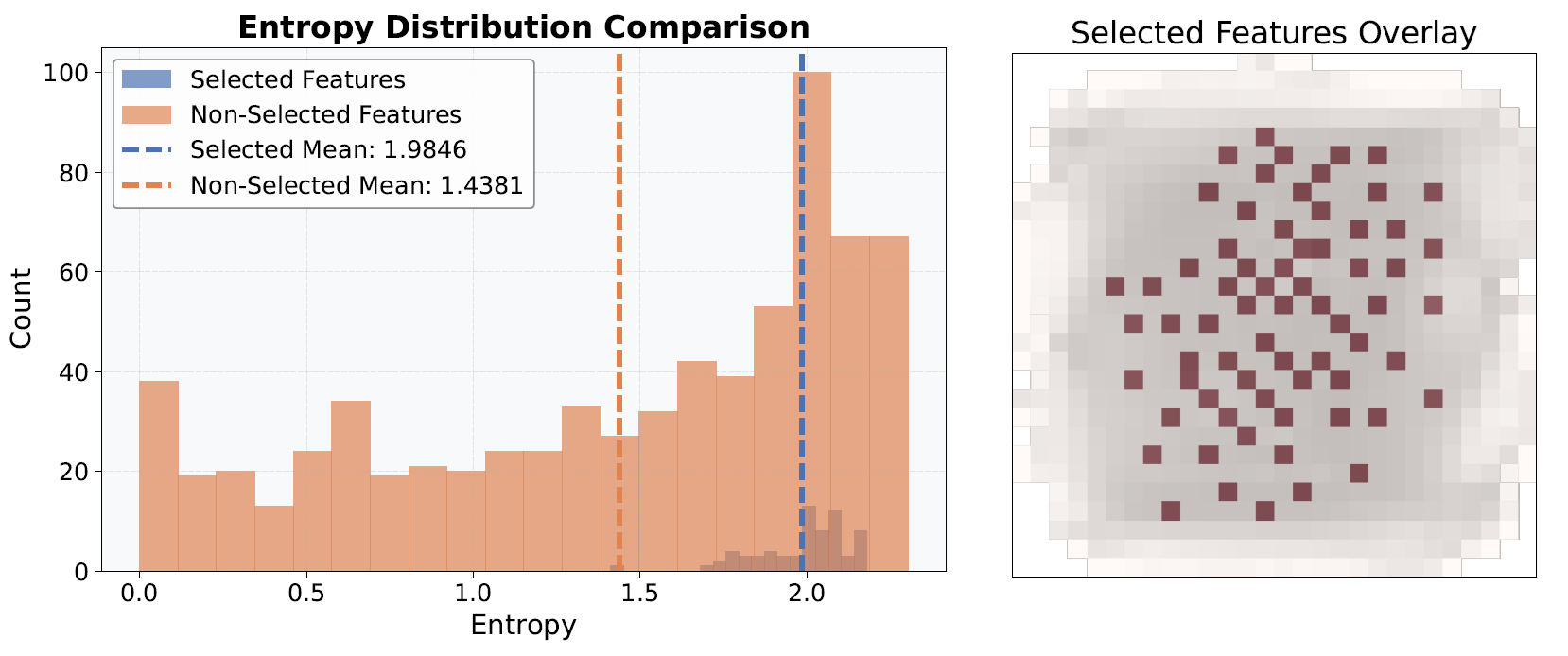}
    \caption{Average entropy of selected features is significantly higher than the entropy of all features, which means that \ours{} selected features with discriminative potential (left). Moreover, selected pixel are localized in the center region of the image (right).}
    \label{fig:avg_entropy}
\end{figure}

\subsection{Feature Selection Visualization on MNIST} \label{app:mnist}

To provide intuitive insights into how \ours{} selects discriminative features, we conducted visualization experiments on the MNIST handwritten digit dataset. \Cref{fig:avg_entropy} (left) compares the entropy of the selected vs. non-selected features. It is evident that \ours{} focuses more on discriminative features (with higher entropy). The mean entropy of selected features equals 1.98 while the mean entropy of all remaining features equals 1.43. \Cref{fig:avg_entropy} (right) illustrates the pixels selected for the MNIST dataset. Clearly, the model pays more attention to the center of the image, ignoring the background regions. \Cref{sec:mnist2} presents more detailed results of this experiment.

% entropy-based feature selection process, showing that selected features demonstrate significantly lower entropy than non-selected ones ($p < 0.001$ in Monte Carlo simulation). The entropy map reveals that pixels near the center of the image generally have lower entropy, and \ours{} predominantly selects features from these regions.

% \Cref{fig:entropy_comparison} (right)  examines individual selected features for a sample digit, comparing their class-conditional activation distributions with non-selected pixels. Selected features consistently show more discriminative patterns across digit classes, with lower entropy values indicating higher information content.

% These visualizations demonstrate that \ours{} selects features in a manner that aligns with human intuition about discriminative regions for digit recognition. Our random sampling analysis confirms that the observed entropy differences between selected and non-selected features represent statistically significant patterns, not merely artifacts of subsampling.

\section{Conclusion}

We presented \ours{}, a novel neural architecture for FS in a differentiable end-to-end manner using temperature-controlled Gumbel-Sigmoid sampling. The key innovation lies in its ability to automatically determine not only which features are relevant but also how many features should be retained, a common pain point in traditional FS methods. Whereas most existing techniques require the number of selected features to be manually specified or found through expensive hyperparameter tuning, \ours{} learns this quantity during training. %The number of features is implicitly controlled by a differentiable regularization term, allowing the model to adaptively select the smallest sufficient subset required to solve the task. This mechanism eliminates the need for retraining across different feature budgets and greatly simplifies the model selection pipeline.

Experimental results in synthetic benchmarks and real-world datasets demonstrate that \ours{} consistently selects fewer features than baselines, without compromising predictive performance. This reduction is beneficial in terms of computational efficiency and interpretability, but also validates the model's ability to avoid overfitting by ignoring redundant or noisy inputs.

% The near-constant computational overhead ($\alpha \approx 0.08$) further distinguishes \ours{} from conventional FS approaches whose cost grows linearly or superlinearly with input dimensionality. In high-dimensional settings--such as metagenomics, where feature counts routinely exceed hundreds--this property makes \ours{} especially practical and scalable.

% In addition to its adaptive feature count, \ours{} produces stable, task-optimized feature masks that remain consistent across samples. This enables robust downstream interpretation and potential reuse in pipeline applications. As shown in our MNIST and metagenomic experiments, the selected features often align with domain-relevant patterns, supporting the case for \ours{} as both a performant and interpretable selector.

Looking ahead, this automatic feature count discovery opens doors for broader applications, such as real-time model compression, adaptive inference, or integration with AutoML frameworks. Moreover, the balance between sparsity and accuracy, controlled through a single $\lambda$ parameter, makes \ours{} a drop-in replacement for feature selectors in a wide range of tasks.

\section{Acknowledgments}
The work of M. \'Smieja was supported by the National Science Center (Poland), grant no. 2023/50/E/ST6/00169.
We gratefully acknowledge Polish high-performance computing infrastructure PLGrid (HPC Center: ACK Cyfronet AGH) for providing computer facilities and support within computational grant no. PLG/2025/018832. The work of Witold Wydma\'nski is supported by the Ministry of Science grant no. PN/01/0195/2022.

\newpage

\bibliographystyle{siamplain}
\bibliography{example_references}

%%%%%%%%%%%%%%%%%%%%%%%%%%%%%%%%%%%

\newpage

\appendix

\section{Details of the benchmark}

\Cref{tab:stats} presents the summary of datasets. \Cref{tab:random_features,tab:corrupted_features,tab:secondorder_features} presents detailed results of the experiments, which are summarized in \Cref{fig:justavg}.

\begingroup

\begin{table}[ht]
\centering
\scriptsize
% \tiny
\setlength{\tabcolsep}{3pt}
\caption{Summary of datasets.}
\begin{tabular}{l|ccc}
\toprule
\textbf{Dataset} & \textbf{Samples} & \textbf{Classes} & \textbf{Features}\\
\midrule
AL (aloi) & 108 000 & 1000 & 128 \\
CH (california) & 20 640 & regression & 8 \\
EY (eye) & 10 936 & 3 & 26 \\
GE (gesture) & 9 873 & 5 & 32 \\
HE (helena) & 65 196 & 100 & 27\\
HI (higgs\_small) & 98 050 & 2 & 28 \\
HO (house) & 22 784 & regression & 16 \\
JA (jannis) & 83 733 & 4 & 54  \\
MI (microsoft) & 1 200 192 & regression & 136  \\
OT (otto) & 61 878 & 9 & 93 \\
YE (year) & 515 345 & regression & 90 \\
\bottomrule
\end{tabular}
\label{tab:stats}
\end{table}

\setlength{\tabcolsep}{2.5pt}

\begin{table*}[htb]
    \centering
    \caption{Classification (accuracy) and regression (negative MSE) performance in the case of random features. Higher values denote better scores.}
    \scriptsize
    % \scalebox{0.9}{
        \begin{tabular}{lllllllllllll}
        \toprule
        Method & AL & CH & EY & GE & HE & HI & HO & JA & MI & OT & YE & rank \\
        \midrule
        No FS & 0.941 & -0.480 & 0.538 & 0.466 & 0.366 & 0.798 & -0.622 & 0.703 & -0.911 & 0.773 & -0.801 & 12.4 \\
        Univariate & \textbf{0.960} & -0.447 & 0.575 & 0.515 & 0.379 & 0.811 & \textbf{-0.549} & 0.715 & \textbf{-0.891} & 0.808 & -0.776 & 4.6 \\
        Lasso & 0.949 & -0.454 & 0.547 & 0.458 & 0.38 & 0.812 & -0.599 & 0.715 & -0.907 & 0.805 & -0.787 & 9.2 \\
        1L Lasso & 0.952 & -0.451 & 0.564 & 0.474 & 0.375 & 0.811 & -0.568 & 0.715 & -0.897 & 0.796 & \textbf{-0.773} & 7.9 \\
        AGL & 0.958 & -0.512 & 0.578 & 0.473 & \textbf{0.386} & 0.81 & -0.557 & 0.718 & -0.898 & 0.799 & -0.778 & 6.9 \\
        LassoNet & 0.954 & -0.445 & 0.552 & 0.495 & 0.385 & 0.811 & -0.557 & 0.715 & -0.907 & 0.783 & -0.787 & 7.7 \\
        AM & 0.953 & -0.444 & 0.554 & 0.498 & 0.382 & 0.813 & -0.566 & 0.722 & -0.904 & 0.801 & -0.777 & 6.0 \\
        RF & 0.955 & -0.453 & 0.589 & \textbf{0.594} & \textbf{0.386} & 0.814 & -0.572 & 0.72 & -0.904 & 0.806 & -0.786 & 5.1 \\
        XGBoost & 0.956 & -0.444 & 0.590 & 0.502 & 0.385 & 0.812 & -0.560 & 0.72 & -0.893 & 0.805 & -0.777 & 4.4 \\
        Deep Lasso & 0.959 & -0.443 & 0.573 & 0.485 & 0.383 & 0.814 & \textbf{-0.549} & 0.72 & -0.894 & 0.802 & -0.776 & 4.3 \\
        STG & 0.940 & -0.500 & 0.533 & 0.478 & 0.371 & 0.8 & -0.630 & 0.701 & OOM & 0.772 & OOM & 12.9 \\
        ProtoGate & 0.922 & N/A & \textbf{0.648} & 0.474 & 0.377 & 0.809 & N/A & 0.724 & N/A & 0.784 & N/A & 10.3 \\
        LSpin & 0.949 & -0.443 & 0.535 & 0.55 & 0.361 & 0.801 & -0.599 & 0.702 & -0.908 & 0.781 & -0.799 & 10.2 \\
        \ours{} & \textbf{0.960} & \textbf{-0.441} & 0.634 & 0.55 & 0.375 & \textbf{0.818} & -0.565 & \textbf{0.738} & -0.893 & \textbf{0.811} & -0.782 & \textbf{3.3} \\
        \bottomrule
        \end{tabular}
    % }
    \label{tab:random_features}
\end{table*}

\begin{table*}[htb]
    \scriptsize
    \centering
    \caption{Classification (accuracy) and regression (negative MSE) performance in the case of corrupted features. Higher values denote better scores.}
    % \scalebox{0.9}{
        \begin{tabular}{lllllllllllll}
        \toprule
         & AL & CH & EY & GE & HE & HI & HO & JA & MI & OT & YE & rank \\
        FS method &  &  &  &  &  &  &  &  &  &  &  &  \\
        \midrule
        No FS & 0.946 & -0.475 & 0.557 & 0.525 & 0.37 & 0.802 & -0.607 & 0.703 & -0.909 & 0.778 & -0.797 & 11.7 \\
        Univariate & 0.955 & -0.451 & 0.556 & 0.514 & 0.346 & 0.810 & -0.620 & 0.717 & -0.920 & 0.795 & -0.828 & 10.3 \\
        Lasso & 0.955 & -0.449 & 0.548 & 0.512 & 0.382 & 0.813 & -0.602 & 0.713 & -0.903 & 0.796 & -0.795 & 8.0 \\
        1L Lasso & 0.955 & -0.447 & 0.566 & 0.515 & 0.382 & 0.812 & -0.581 & 0.718 & -0.902 & 0.795 & -0.780 & 6.7 \\
        AGL & 0.953 & -0.450 & 0.588 & 0.538 & 0.386 & 0.813 & -0.561 & 0.722 & -0.902 & 0.796 & -0.780 & 5.0 \\
        LassoNet & 0.955 & -0.452 & 0.57 & 0.556 & 0.382 & 0.811 & -0.551 & 0.719 & -0.905 & 0.795 & -0.777 & 6.0 \\
        AM & 0.955 & -0.449 & 0.583 & 0.527 & 0.381 & 0.814 & -0.555 & 0.722 & -0.905 & 0.797 & -0.780 & 5.2 \\
        RF & 0.951 & -0.453 & 0.574 & 0.568 & 0.383 & 0.810 & -0.565 & 0.724 & -0.904 & 0.788 & -0.786 & 6.9 \\
        XGBoost & 0.954 & -0.454 & 0.583 & 0.51 & 0.385 & 0.815 & -0.553 & 0.722 & \textbf{-0.892} & 0.803 & -0.779 & 5.0 \\
        Deep Lasso & 0.955 & -0.447 & 0.577 & 0.525 & \textbf{0.388} & 0.815 & -0.567 & 0.721 & -0.895 & 0.801 & \textbf{-0.776} & 3.9 \\
        STG & 0.945 & -0.475 & 0.559 & 0.523 & 0.371 & 0.805 & -0.589 & 0.706 & OOM & 0.782 & OOM & 11.6 \\
        ProtoGate & 0.918$^*$ & N/A & 0.54 & 0.525 & 0.379 & 0.802$^*$ & N/A & 0.711$^*$ & N/A & 0.781 & N/A & 12.5 \\
        LSpin & 0.947$^*$ & -0.449 & 0.554 & \textbf{0.575} & 0.366 & 0.802$^*$ & -0.611 & 0.707$^*$ & -0.907$^*$ & 0.78 & -0.796$^*$ & 10.2 \\
        \ours{} & \textbf{0.957} & \textbf{-0.437} & \textbf{0.625} & 0.57 & 0.373 & \textbf{0.819} & \textbf{-0.549} & \textbf{0.735} & -0.895 & \textbf{0.804} & -0.779 & \textbf{2.3} \\
        \bottomrule
        \end{tabular}
    % }
    \label{tab:corrupted_features}
\end{table*}

\begin{table*}[htb]
    \caption{Classification (accuracy) and regression (negative MSE) performance in the case of second-order features. Higher values denote better scores.}
    \scriptsize
    \centering
    % \scalebox{0.9}{
        \begin{tabular}{lllllllllllll}
        \toprule
         & AL & CH & EY & GE & HE & HI & HO & JA & MI & OT & YE & rank \\
        FS method &  &  &  &  &  &  &  &  &  &  &  &  \\
        \midrule
        No FS & 0.960 & -0.443 & 0.631 & 0.605 & 0.383 & 0.811 & -0.549 & 0.719 & -0.891 & 0.800 & -0.786 & 8.1 \\
        Univariate & \textbf{0.961} & -0.439 & 0.584 & 0.582 & 0.357 & 0.817 & -0.614 & 0.724 & -0.902 & 0.798 & -0.810 & 10.0 \\
        Lasso & 0.955 & -0.443 & 0.608 & 0.59 & 0.366 & 0.816 & -0.564 & 0.724 & -0.891 & 0.806 & -0.783 & 8.8 \\
        1L Lasso & 0.959 & -0.445 & 0.634 & 0.571 & 0.380 & 0.815 & -0.565 & 0.728 & \textbf{-0.890} & \textbf{0.808} & -0.780 & 7.4 \\
        AGL & \textbf{0.961} & -0.443 & 0.637 & 0.594 & 0.383 & 0.807 & -0.565 & 0.730 & \textbf{-0.890} & 0.806 & -0.776 & 6.3 \\
        LassoNet & 0.959 & -0.442 & 0.641 & 0.611 & 0.379 & 0.816 & -0.595 & 0.724 & -0.893 & 0.797 & -0.784 & 7.9 \\
        AM & \textbf{0.961} & -0.439 & 0.622 & 0.604 & 0.381 & \textbf{0.819} & -0.566 & 0.730 & -0.892 & 0.802 & -0.778 & 6.1 \\
        RF & 0.958 & -0.437 & 0.639 & \textbf{0.619} & 0.370 & 0.818 & -0.586 & 0.735 & \textbf{-0.890} & 0.801 & -0.781 & 5.6 \\
        XGBoost & 0.870 & -0.438 & 0.635 & 0.604 & 0.373 & 0.818 & -0.579 & 0.734 & -0.891 & 0.805 & -0.786 & 7.4 \\
        Deep Lasso & \textbf{0.961} & -0.441 & \textbf{0.648} & 0.6 & 0.384 & 0.815 & -0.572 & 0.733 & \textbf{-0.890} & 0.805 & -0.776 & 5.0 \\
        STG & 0.960 & -0.443 & 0.616 & 0.609 & \textbf{0.389} & 0.810 & \textbf{-0.540} & 0.719 & OOM & 0.800 & OOM & 8.7 \\
        ProtoGate & 0.922$^*$ & N/A & 0.64 & 0.615 & 0.374$^*$ & 0.808$^*$ & N/A & 0.711$^*$ & N/A & 0.796$^*$ & N/A & 11.4 \\
        LSpin & 0.954$^*$ & -0.450 & 0.637 & 0.608 & \textbf{0.389}$^*$ & 0.809$^*$ & -0.547 & 0.723$^*$ & -0.895$^*$ & 0.806 & -0.793 & 8.1 \\
        \ours{} & 0.960 & \textbf{-0.436} & 0.638 & 0.6 & 0.378 & 0.817 & -0.548 & \textbf{0.738} & -0.891 & \textbf{0.808} & \textbf{-0.775} & \textbf{4.4} \\
        \bottomrule
        \end{tabular}
    % }
    \label{tab:secondorder_features}
\end{table*}

\endgroup

\section{Details of visualization on MNIST dataset}
\label{sec:mnist2}

\begin{figure}[h]
    \centering    \includegraphics[width=\linewidth]{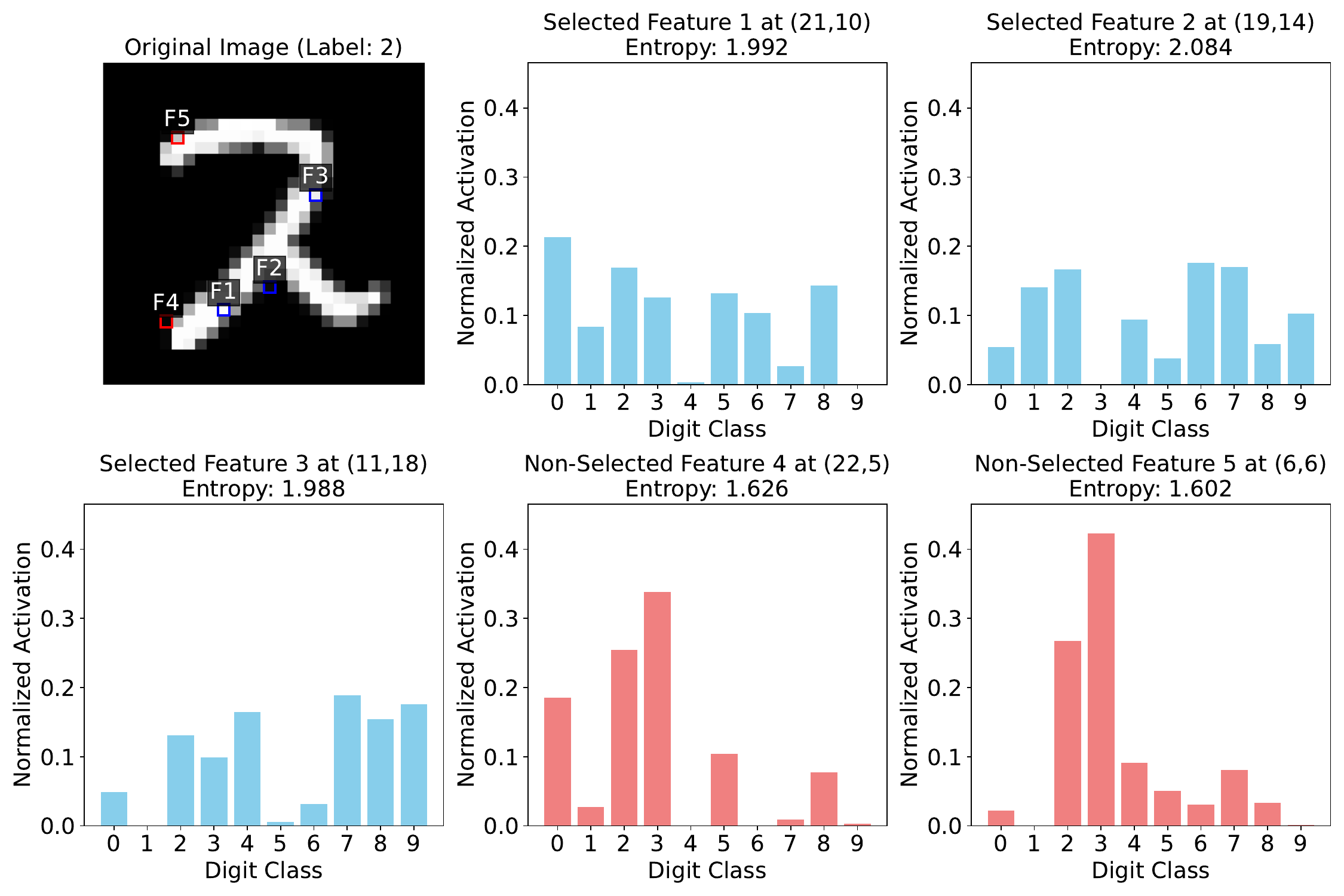}
    \caption{Analysis of sample features (top-left) from MNIST dataset shows that entropy of selected features (F1-F3) is much higher than their non-selected counterparts (F4, F5). It confirms that \ours{} selects the most discriminative features.}
    \label{fig:entropy_comparison}
\end{figure}

We continue visualization experiments on the MNIST handwritten digit dataset (see \Cref{app:mnist}). \Cref{fig:entropy_comparison} examines individual selected features (blue) for a sample digit, comparing their class-conditional activation distributions with non-selected pixels (red). Selected features consistently show more discriminative patterns across digit classes, with higher entropy values indicating higher information content. These visualizations demonstrate that \ours{} selects features in a manner that aligns with human intuition about discriminative regions for digit recognition. 

% SIAM recommends using BibTeX
% if using BibTeX
\end{document}